\documentclass[11pt,a4paper]{article}
\usepackage[hyperref]{conll-2019}
\usepackage{times}
\usepackage{latexsym}
\usepackage{booktabs}
\usepackage{url}
\usepackage{multirow}
\usepackage{caption}
\usepackage{subcaption}
\usepackage{graphicx}
\usepackage{xcolor}
\usepackage{wrapfig, framed}
\usepackage{xspace}

\aclfinalcopy

\title{A Case Study on Combining ASR and Visual Features\\for Generating Instructional Video Captions}

\author{
  \begin{tabular}{cc}
    Jack Hessel & \hspace{.31in} Bo Pang \quad Zhenhai Zhu \quad Radu Soricut \\
    {\normalfont Cornell University} & {\normalfont Google} \\
    {\tt jhessel@cs.cornell.edu} & {\tt \{bopang,zhenhai,rsoricut\}@google.com} \\
  \end{tabular}
}

\date{}

\newcommand{\mparagraph}[1]{\noindent\textbf{{#1}}.}
\newcommand{\mparagraphnp}[1]{\noindent\textbf{{#1}}}

\definecolor{goodperformance}{HTML}{38761d}
\definecolor{okayperformance}{HTML}{e69138}
\definecolor{badperformance}{HTML}{cc0000}

\newcommand{\aucmath}{{\scriptstyle \sf{AUC}}}
\newcommand{\auc}{$\aucmath$\xspace}
\newcommand{\aucmean}{$\aucmath_{\mu,w}$\xspace}
\newcommand{\aucmeanw}[1]{$\aucmath_{\mu,#1}$\xspace}
\newcommand{\aucdelta}{$\aucmath_{\Delta,w}$\xspace}
\newcommand{\aucdeltaw}[1]{$\aucmath_{\Delta,#1}$\xspace}
\newcommand{\aucspeech}{$\aucmath_{t,w}$\xspace}
\newcommand{\aucvideo}{$\aucmath_{v,w}$\xspace}
\newcommand{\aucspeechw}[1]{$\aucmath_{t,#1}$\xspace}
\newcommand{\aucvideow}[1]{$\aucmath_{v,#1}$\xspace}
\newcommand{\metric}[1]{{\small #1}\xspace}
\newcommand{\bleu}{\metric{BLEU-4}}
\newcommand{\meteor}{\metric{METEOR}}
\newcommand{\rouge}{\metric{ROUGE-L}}
\newcommand{\cider}{\metric{CIDEr}}

\newcommand{\tinymetric}[1]{{\scriptsize #1}\xspace}

\newcommand{\tinybleu}{\tinymetric{BLEU-4}}
\newcommand{\tinymeteor}{\tinymetric{METEOR}}
\newcommand{\tinyrouge}{\tinymetric{ROUGE-L}}
\newcommand{\tinycider}{\tinymetric{CIDEr}}

\definecolor{hard}{HTML}{e66101}
\definecolor{easy}{HTML}{5e3c99}
\definecolor{vid_better}{HTML}{4dac26}
\definecolor{speech_better}{HTML}{d01c8b}

\begin{document}
\maketitle

\begin{abstract}
Instructional videos get high-traffic on video sharing platforms, and prior work suggests that providing time-stamped, subtask annotations (e.g., ``heat the oil in the pan") improves user experiences. However, current automatic annotation methods based on visual features alone perform only slightly
better than constant prediction.
Taking cues from prior work, we show that we can improve performance significantly by considering automatic speech recognition (ASR) tokens as input. Furthermore, jointly modeling ASR tokens and visual features results in higher performance compared to training individually on either modality.
We find that unstated background information is better explained by visual features, whereas fine-grained distinctions (e.g., ``add oil" vs. ``add olive oil") are disambiguated more easily via ASR tokens.
\end{abstract}

\section{Introduction}

Instructional videos increasingly dominate user attention on online video platforms. For example, 86\% of YouTube users report using the platform often to learn new things, and 70\% of users report using videos to solve problems related to work, school, or hobbies \cite{oneilhart2018}.

Prior work in user experience has investigated the best way of presenting instructional videos to users.
Kim et al.~\shortcite{kim2014crowdsourcing}, for example, compare two options; first: presenting users with the video alone, and second: presenting the video with an additional \emph{structured} representation, including a timeline populated with task subgoals.
Users interacting with the structured video representation reported higher satisfaction, and external judges rated
the work they completed using the videos as having higher quality.
Margulieux et al.~\shortcite{margulieux2012subgoal} and Weir et al.~\shortcite{weir2015learnersourcing} similarly find that presenting explicit subgoals alongside how-to videos improves user experiences.
Thus, presenting instructional videos with additional structured annotations is likely to benefit users.

\begin{figure}[t]
    \centering
    \includegraphics[width=.8\linewidth]{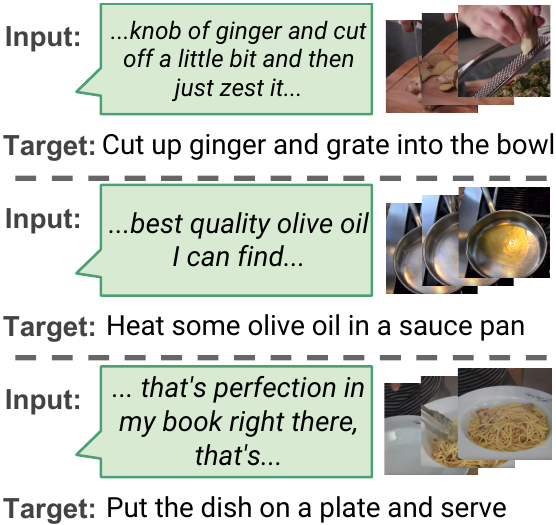}
    \caption{Illustration of a multimodal dense instructional video captioning task. Models are given access to both video frames and ASR tokens, and must generate a recipe instruction step for each video segment. The speaker in the video \emph{sometimes} (but not always) references literal objects and actions.}
    \label{fig:example_datapoints}
\end{figure}

These studies rely on human annotation of time-stamped subtask goals, e.g., timed captions created through crowdsourcing. However, human-in-the-loop annotation is infeasible to deploy for popular video sharing platforms like YouTube that receive hundreds of hours of uploads per minute.
In this work, we address the task of \emph{automatically} producing captions for instructional videos at the level of video segments.
Ideally, generated captions provide a literal, imperative description of the procedural step occurring for a given video segment, e.g., in the cooking context we consider, ``add the oil to the pan."

Producing segment-level captions is a sub-task of dense video captioning, 
where prior work has mostly focused on visual-only models. 
Dense captioning is a difficult task, particularly in the instructional video domain, as
fine-grained distinctions may be difficult or impossible to make with visual features alone. 
Visual information can be ambiguous (e.g., distinguishing between ``olive oil" vs. ``vegetable oil") or incomplete (e.g., preparation steps may occur off-camera). In our study, a first important finding is that, for the dataset considered, current state-of-the-art, visual-features--only models only slightly outperform a constant prediction baseline, e.g., by 1.5 BLEU/METEOR points.

To improve performance in this difficult setting, we consider the \emph{automatic speech recognition} (ASR) tokens generated by YouTube. These publicly available tokens are an ASR model's attempts to map words spoken in videos into text. However, while a promising potential source for signal, it is not always trivial to transform even accurate ASR into the desired imperative target: while there are cases of clear correspondence between the literal actions in the video and the ASR tokens, in other cases, the mapping is imperfect (Fig.~\ref{fig:example_datapoints}). For example, when finishing a dish, a user says ``that's perfection in my book right there" rather than ``put the dish on a plate and serve." There are also cases where no ASR tokens are available at all. Despite these potential difficulties, previous work has demonstrated that ASR can be informative in a variety of instructional video understanding tasks~\cite{naim2014unsupervised,naim2015discriminative,malmaud2015s,sener2015unsupervised,alayrac2016unsupervised,huang2017unsupervised};
though less work has focused on instructional caption \emph{generation,} which is known to be difficult and sensitive to input perturbations \cite{chen2018attacking}.

We find that incorporating ASR-token--based features significantly improves performance over visual-features--only models
(e.g., CIDEr improves 0.53 $\Rightarrow$ 1.0, BLEU-4 improves 4.3 $\Rightarrow$ 8.5).
We also show that \emph{combining} ASR tokens and visual features results in the highest performing models,
suggesting that the modalities contain complementary information.

We conclude by asking: what information is captured by the visual features that \emph{is not} captured by the ASR tokens (and vice versa)?
Auxiliary experiments examining performance of models in predicting the presence/absence of individual word types suggest that visual signals are superior for identifying unspoken, implicit aspects of scenes; for instance, in order to mix ingredients, they must be placed in a bowl --- and although bowls are often visually present in the scene, ``bowl" is often not explicitly mentioned by the speaker.
Conversely, 
ASR features
readily disambiguate between fine-grained entities, e.g., ``olive oil" vs.``vegetable oil", a task that is difficult (and sometimes impossible) for visual features alone.

\section{Related Work}

\mparagraph{Narrated instructional videos}
While several works have matched audio and video signals in an unconstrained setting \cite{arandjelovic2017look,tian2018audio}, our work builds upon previous efforts to utilize accompanying speech signals to understand online \emph{instructional} videos, specifically.
Several works focus on learning video-instruction alignments, and match a fixed set of instructions to temporal video segments \cite{regneri2013grounding,naim2015discriminative,malmaud2015s,hendricks2017localizing,kuehne2017weakly}. Another line of previous work uses speech to extract and align language fragments, e.g., verb-noun pairs, with instructional videos \cite{gupta2010using,motwani2012improving,alayrac2016unsupervised,huang2017unsupervised,huang2018finding,hahn2018learning}.
\newcite{sener2015unsupervised}, as part of their parsing pipeline, train a 3-gram language model on segmented ASR token inputs to produce recipe steps.

\mparagraph{Dense Video Captioning}
Recent work in computer vision addresses dense video captioning \cite{krishna2017dense,li2018jointly,wang2018bidirectional}, a supervised task that involves (i) segmenting the input video, and, (ii) generating a natural language description for each segment.
Here, we focus on the second subtask of generating descriptions given a ground-truth segmentation; this setting isolates the language generation part of the modeling process.\footnote{We find that state-of-the-art models perform poorly even for just this subtask (see \S~\ref{sec:data:baseline}), so we reserve the full task for future work.} 
Most related to the present work are several dense captioning approaches that have been applied to instructional videos \cite{zhou2017towards,zhou2018end}.
\newcite{zhou2018end} achieve state-of-the-art performance on the dataset we consider; their model is video-only, and combines a region proposal network~\cite{ren2015faster} and a Transformer~\cite{vaswani2017attention} decoder.

\mparagraph{Multimodal Video Captioning} Several works have employed multimodal signals to caption the MSR-VTT dataset \cite{xu2016msr}, which consists of 2K video clips from 20 general categories (e.g., ``news", ``sports") with an average duration of 10 seconds per clip. In particular, \newcite{ramanishka2016multimodal,xu2017learning,hori2017attention,shen2017weakly,chuang2017seeing,hao2018integrating} all report small performance gains when incorporating audio features on top of visual features. However --- we suspect that instructional video domain is significantly different than MSR-VTT (where the audio information does not necessarily correspond to human speech), as we find that ASR-only models significantly surpass the state-of-the-art video model in our case. \newcite{palaskar2019multimodal} and \newcite{shi2019dense}, contemporaneous with the submission of the present work, also examine ASR as a source of signal for generating how-to video captions.

\section{Dataset}
We focus on YouCook2 ~\cite{zhou2017towards}, the largest human-captioned dataset of instructional videos publicly available.\footnote{How2 \cite{sanabria2018how2} tackles the different task of predicting video uploader-provided descriptions/captions, which are not always appropriate summarizations.}
It contains 2000 YouTube cooking videos, for a total of 176 hours, and spans 89 different recipes.
Each video averages at 5.26 minutes, and is annotated with an average of 7.7 temporal segments (i.e., start/end points) corresponding to semantically distinct recipe steps.
Each segment is associated with an imperative caption, e.g., ``add the oil to the pan", for an average of 8.8 words per caption.

At the time of analysis (June 2018), over 25\% of the YouCook2 videos had been removed from YouTube, and therefore we do not consider them.
As a result, all our experiments operate on a \emph{subset} of the YouCook2 data.
While this makes direct comparison with previous and future work more difficult,
our performance metrics can be viewed as lower bounds, as they are trained on less data compared to, e.g., ~\cite{zhou2018end}. Unless noted otherwise, our analyses are conducted over 1.4K videos and the 10.6K annotated segments contained therein.

\subsection{A Closer Look at ASR tokens}
\label{sec:data:asr}

We collected the ASR tokens automatically generated by YouTube (available through the YouTube Data API\footnote{https://developers.google.com\/youtube/v3/docs/captions} with trackKind = ASR), which are then mapped to their temporally corresponding video segments. We start by asking the following questions: How much narration do users provide for instructional videos? And: can YouTube's ASR system detect that speech?

Not surprisingly, speakers in videos tend to be more verbose than  the annotated groundtruth captions:
we find the length distribution of ASR tokens per segment to be roughly log-normal, with mean/median length being 42/28 tokens respectively (compared to a mean of 9 tokens/segment for captions).  %
Over the 10.6K available segments, only 1.6\% of them have zero associated tokens.
Furthermore, based on automatic language identification provided by the YouTube API and some manual verification, we estimated that less than 1\% of videos contain completely non-English speech (but we do not discard them from our experiments).

We also investigate the words-per-minute (WPM) ratio, based on the video segment length. %
The mean value of 134 WPM is slightly lower than, but comparable to, previously reported figures of English speaking rates~\cite{yuan2006towards}, which indicates that, for this set of video segments, words are being detected at rates comparable to everyday English speech.

\subsection{A Closer Look at the Generation Task}

To better understand the generation task, we computed lower and upper bounds for generation performance using a constant-prediction baseline and human performance, respectively.
\label{sec:data:baseline}

\mparagraph{Lower bound: constant}  For all segments at test time, we predict ``heat some oil in a pan and add salt and pepper to the pan and stir."
This sentence is constructed by examining the most common n-grams in the corpus and pasting them together.

\mparagraph{Upper bound: human estimate} 
We conducted a small-scale experiment to estimate human performance for the segment-level captioning task.
Two of the authors of this paper, after being trained on segment-level captions from three videos, attempted to mirror that style of annotation for the segments of 20 randomly sampled videos, totalling over 140 segment annotations each.\footnote{These preliminary experiments are not meant to provide a definitive, exact measure of inter-annotator agreement.} Both human annotators report low-confidence with the task, in particular, they found it difficult to maintain a consistent level of specificity in terms of how many factual details to include (e.g., ``mix together" vs. ``mix the peppers and mushrooms together.")

\label{sec:sec-with-captioning-metrics}
\mparagraphnp{Results:} We compute corpus-level performance statistics using four standard generation evaluation metrics: \rouge~\cite{lin2004rouge}, \cider~\cite{vedantam2015cider}, \bleu~\cite{papineni2002bleu} and \meteor~\cite{banerjee2005meteor} (higher is better in all cases).

\emph{Note that our evaluation is micro-averaged at the segment level, and differs slightly from prior work on this dataset,} which has mostly reported metrics macro-averaged at the video level. We switched the evaluation because some metrics like \bleu exhibit undesirable sparsity artifacts when macro-averaging, e.g., any video without a correct 4-gram gets a zero BLEU score, even if there are many 1/2/3-grams correct. Segment-level averaging, the standard evaluation practice in fields like machine translation, is insensitive to this sparsity concern, and (we believe) provides a more robust perspective on performance.

\begin{table}[h]
{\footnotesize
\begin{center}
\begin{tabular}{ r  c@{\hspace{.1cm}}c@{\hspace{.1cm}}c@{\hspace{.1cm}}c}
   & \tinybleu & \tinymeteor & \tinyrouge & \tinycider \\ \midrule
 Constant Prediction     & 2.70 & 10.3 & 21.7 & .15 \\ \midrule
\newcite{zhou2018end}  & 3.84 & 11.6 & 27.4 & .38
\\
\newcite{sun2019videobert} & 4.07 & 11.0 & 27.5 & .50
\\
\newcite{sun2019contrastive} & 4.31 & 11.9 & 29.5 & .53
\\ \midrule
 Human Estimate        & 15.2 & 25.9 & 45.1 & 3.8\\
\end{tabular}
\end{center}}
\caption{The performance of several state-of-the-art, video-only models, with lower (constant prediction) and upper (human estimate) bounds.}
\label{tab:const-baseline}
\end{table}

This comparison highlights the gap that remains between the simplest possible baseline, several computer vision based models, and (roughly) how well humans perform at this task. Given that \newcite{sun2019contrastive} is a highly tuned computer vision model transfer learned from a corpus of over 300K cooking videos, from the perspective of building video captioning systems in practice, we suspect that incorporating additional modalities like ASR is more likely to result in performance gains versus building better computer vision models.

\section{Models}
In addition to the constant prediction baseline, 
we explore a series of ASR-based baseline methods:%

\mparagraphnp{ASR as the Caption (ASC)}
This baseline returns the test-time ASR token sequence as the caption. While the result is not a coherent, imperative step, performance of this method offers insight into the extent of word overlap between the ASR sequence and the target groundtruth, as measured by the captioning metrics.

\mparagraphnp{Filtered ASR (FASC)} Given that the ASR token sequences are much longer than groundtruth captions (\S~\ref{sec:data:asr}), the performance of ASC incurs a length (or precision-based) penalty for several metrics.  %
The FASC baseline strengthens ASC by removing 
word types that are less likely to appear in groundtruth captions,
e.g., ``ah'', ``he'', ``hello," or ``wish''.
Specifically, we only keep words with high $\frac{P(w ~|~ GT)}{P(w ~|~ ASR)}$ values, i.e., words that would be indicative of the groundtruth class if we were to build a Naive-Bayes classifier with add-one smoothing; probabilities are computed only over the training set to reduce the risk of overfitting.
This baseline produces outputs that are shorter compared to ASC, but it is unlikely to yield fluent, readable text.

\mparagraphnp{ASR-based Retrieval (RET)}
This retrieval baseline memorizes the recipe steps in the training set, and represents them each as tf-idf vectors.
At test-time, the ASR sequence is converted into a tf-idf vector and compared to each training-set caption via cosine similarity.\footnote{We tried several variants of this method, e.g., comparing test ASR to train ASR, but found that comparing test ASR to train captions performed the best.} 
The training caption that is most similar to the test-time ASR according to this metric is returned as the ``generated" caption.
Note that, although a memorization-based technique, this baseline method produces de-facto captions as outputs.

\subsection{Transformer-based Neural Models}
We explore neural encoder-decoder models based on Transformer Networks~\cite{vaswani2017attention}. In contrast to RNNs, Transformers abandon recurrence in favor of a mix of different types of feed-forward layers, e.g., in the case of the Transformer decoder,
self-attention layers, cross-attention layers (attending to the encoder outputs), and fully connected feed-forward layers. We explore two variants of the Transformer, corresponding to different hypotheses about what information might be useful for captioning instructional videos.

\mparagraphnp{ASR Transformer (AT)}
This model learns to map ASR-token sequences directly to captions using a standard sequence-to-sequence Transformer architecture. The model's parameters are optimized to maximize the probability of the ground-truth instructions, conditioned on the input ASR sequences.

\begin{figure}[t]
\centering
\includegraphics[width=0.4\textwidth]{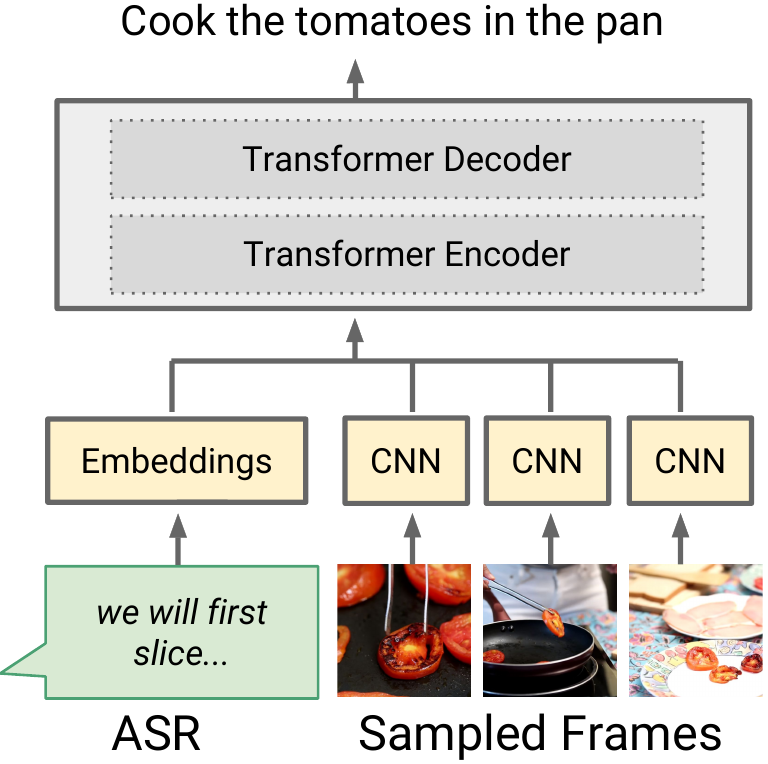}
\caption{The AT+Video model. Both the encoder and decoder layers perform cross-modal attention.}
\label{fig:multimodal_model}
\end{figure}

\mparagraphnp{Multimodal model (AT+Video)}
We incorporate video features into the ASR transformer (Fig~\ref{fig:multimodal_model}).  For ease of comparison with prior and future work, we use features extracted from ResNet34 \cite{he2016deep} pretrained on the ImageNet classification task; these features are provided in the YouCook2 data release. Each video is initially uniformly sampled at 512 frames, with an average of 30 frames per captioned-segment.

To represent each video segment, first, $k$ frames are randomly sampled with replacement. The sampled frames are temporally sorted to preserve ordering information, and their corresponding ResNet34 feature vectors are projected to the Transformer encoder hidden dimension via a width-1 1D convolution. We use $k=10$ for all our experiments. The encoder self-attention layers perform \emph{cross-modal attention} operations between the visual features and the ASR-token--based features. For each output token, the decoder attends to previously predicted tokens, and encoder outputs for all input frames~/~ASR tokens.

\begin{table}
    \centering

\footnotesize
\begin{tabular}{l@{\hspace{.12cm}}c@{\hspace{.12cm}}c@{\hspace{.12cm}}c@{\hspace{.12cm}}c}
 & \bleu & \meteor & \rouge & \cider \\
 \midrule
 CNST & 2.70 & 10.03 & 21.69 & 0.15 \\
 \newcite{sun2019contrastive} & 4.31 & 11.91 & 29.47 & 0.53
 \\
 \midrule
ASC & 1.68 & 14.86 & 19.24 & 0.20 \\
FASC & 4.32 & \underline{18.47} & 30.07 & 0.59 \\
RET & 5.68 & 14.29 & 28.06 & 0.80 \\
AT & \underline{8.55} & 16.93 & 35.54 & 1.06 \\
\midrule
AT+Video & \underline{9.01} & \underline{17.77} & \textbf{36.65} & \textbf{1.12} \\
\end{tabular}

    \caption{Caption generation performance: AT+Video is a multimodal model that adds visual frame features to AT. A bolded value in a column indicates a statistically-significant improvement, whereas an underline indicates a statistical tie for best ($p<.01$).} 
    \label{tab:main_results}
\end{table}

\begin{figure*}
    \centering
    \includegraphics[width=\linewidth]{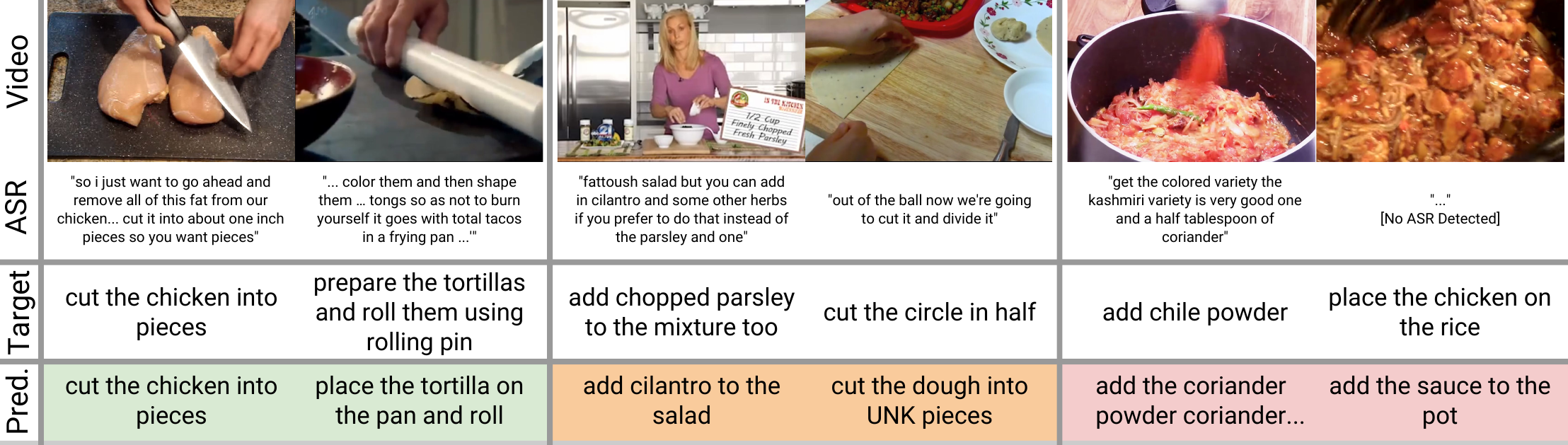}
    \caption{Example generations from AT+Video in cases where it performs
    \textbf{\textcolor{goodperformance}{well}},
    \textbf{\textcolor{okayperformance}{okay}},
    and \textbf{\textcolor{badperformance}{poorly}}.}
    \label{fig:example_outputs}
\end{figure*}

\section{Experiments}
\label{sec:experiments}
We perform 10-fold cross-validation with randomly sampled 80/10/10 train/dev/test splits (split at the video-level), using the same splits for all models.
After discarding the videos that were deleted at the time of data collection, each split contains roughly 1.1K training videos (averaging 8.3K training segments). We report mean performance over these splits according to four standard captioning accuracy metrics, introduced in \S\ref{sec:sec-with-captioning-metrics}. \rouge, \cider, \bleu, and \meteor.
We perform both Wilcoxon signed-rank tests \cite{demvsar2006statistical} and two-sided corrected resampled t-tests \cite{nadeau2000inference} to estimate statistical significance.
To be conservative and reduce the chance of Type I error, we take whichever $p$-value is larger between these two tests.

\mparagraph{Transformer-based model details}
For each cross-validation split,  we use a batch size of 128, tie the Transformer model's feed forward and model dimensions $d_{ffn} = d_{model}$, and optimize regularized cross-entropy loss using Adam \cite{kingma2014adam} with $lr=.001$. We train models for 100K steps, storing checkpoint files periodically.
For each split, we train 8 model variants, conducting a grid search over model dimension, number of encoder/decoder layers, and L2 regularization:
we consider all model parameter settings in $(d_{model}, N_{layer}, \lambda_{reg}) \in \{128,256\} \times \{2,3\} \times \{.0005, .001\}$ for each cross-validation split independently, and select the highest performing, checkpointed model according to \rouge over the development set for that fold.
Transformer models are implemented using \texttt{tensor2tensor} \cite{tensor2tensor} and \texttt{Tensorflow} \cite{tensorflow2015-whitepaper}.
The vocabulary (average size 800) is determined separately using the training data for each cross-validation split.
Words are considered if they occur at least 5 times in the ground-truth of the current training set.\footnote{Different vocabulary creation schemes, e.g., sub-word tokenization, led to small performance decreases.}
This leads to an OOV rate of {\raise.17ex\hbox{$\scriptstyle\sim$}}60\% in the input.
We truncate inputs at 80 tokens ({\raise.17ex\hbox{$\scriptstyle\sim$}}10-15\% of transcripts are truncated in this process).
For simplicity, decoding is done greedily in all cases.

\mparagraph{Generation Experiment Results}
Table~\ref{tab:main_results} reports the performance of each model.
For unimodal models, simple baselines like FASC (filtered ASR) and RET (training-caption retrieval) outperform the state-of-the-art video-only model of \newcite{sun2019contrastive}, according to the four automatic evaluation metrics. Overall, AT yields the best unimodal performance.
Combining ASR and visual signals into a multimodal representation performs even better: the AT+Video model tends to outperform AT (and \newcite{sun2019contrastive}), according to \rouge, \cider, and \meteor ($p$ $<$.01).
Since AT and AT+Video have identical architectures and differ only in the available inputs, this result provides strong evidence that it is indeed the \emph{multimodality} of AT+Video that leads to the (statistically significant) performance gains over the strongest unimodal models.
We present some output examples in Fig.~\ref{fig:example_outputs}.

\subsection{Diversity of Generated Captions}

In addition to the automatic quality metrics, we measure how diverse the generated caption are for each model,
using the following metrics:
vocabulary coverage (the percent of vocabulary that was predicted at test-time by each algorithm at least once); proportion not copied (the percent of generated captions that do not appear in the training set verbatim);
and output uniqueness (the percent of generated captions that are unique).
These metrics are useful because they can highlight undesirable, degenerate behavior for models.\footnote{For instance, the constant prediction baseline we consider would score low in both vocab coverage and uniqueness.}
As an upper-bound, we compute these metrics for the ground-truth (GT) test-time targets. Note that even the ground-truth targets do not achieve 100\% in these diversity metrics: for vocabulary coverage, not all vocabulary items appear in the ground-truth captions for a given cross-validation split; similarly, for proportion not copied/output uniqueness, because there are repeated captions in the label set.

\begin{figure}[ht]
    \centering
    \includegraphics[width=0.7\linewidth]{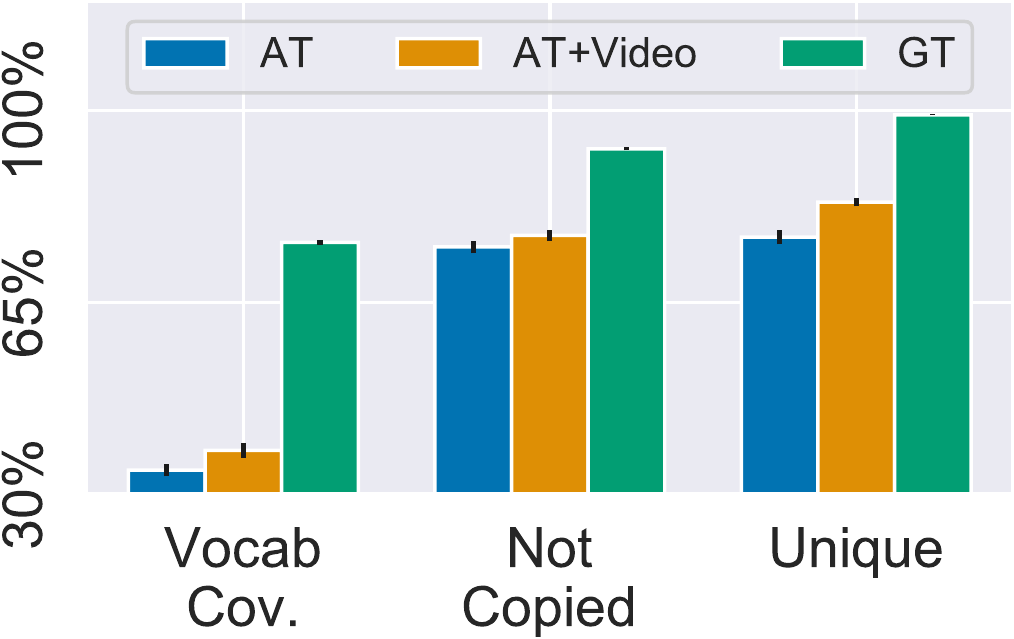}
    \caption{The multimodal model AT+Video produces slightly more diverse captions than its unimodal counterparts.}
    \label{fig:diversity_statistics}
\end{figure}
According to all metrics, AT+Video outputs are slightly more diverse compared to the AT outputs (Fig.~\ref{fig:diversity_statistics}).
This observation suggests that the multimodal model is not simply exploiting a degeneracy to achieve its performance improvements.

\begin{figure*}[ht]
    \begin{subfigure}[t]{0.35\linewidth}
        \centering
        \includegraphics[width=.95\linewidth]{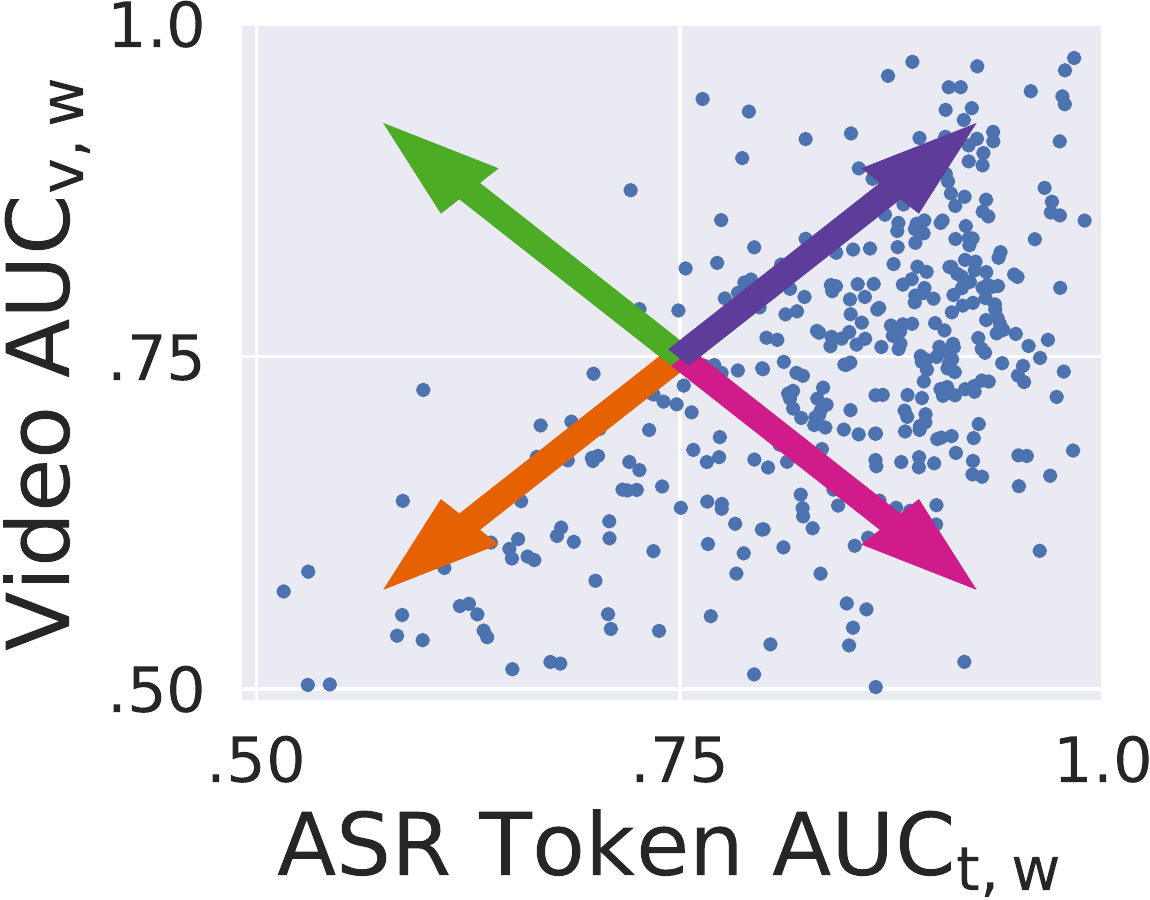}
    \end{subfigure} %
    \begin{subtable}[b]{0.145\linewidth}
        \centering
        \tiny
\begin{tabular}[b]{lc}
\toprule
knead & 97.8 \\
nori & 97.1 \\
yeast & 96.1 \\
mozzarella & 95.8 \\
lettuce & 95.3 \\
pancake & 94.7 \\
wrapper & 94.3 \\
patty & 93.4 \\
dal & 93.0 \\
grill & 92.9 \\
pizza & 92.9 \\
oven & 92.7 \\
bake & 92.3 \\
\bottomrule
\end{tabular}
        \caption{\centering{\textbf{\color{easy}Easiest\newline(\aucmean)}}}
        \label{tab:easiest}
    \end{subtable} %
    \begin{subtable}[b]{0.145\linewidth}
        \centering
        \tiny
\begin{tabular}[b]{lc}
\toprule
4 & 43.8 \\
bit & 51.7 \\
about & 52.1 \\
prepare & 52.3 \\
mixed & 54.5 \\
then & 54.5 \\
spoon & 54.9 \\
or & 55.9 \\
it & 56.2 \\
ready & 56.7 \\
3 & 57.1 \\
few & 58.3 \\
more & 58.5 \\
\bottomrule
\end{tabular}
        \caption{\centering{\textbf{\color{hard}Hardest\newline(\aucmean)}}}
        \label{tab:hardest}
    \end{subtable} %
    \begin{subtable}[b]{0.145\linewidth}
        \centering
        \tiny
\begin{tabular}[b]{lc}
\toprule
fat & 39.7 \\
turn & 36.4 \\
sea & 35.9 \\
white & 31.7 \\
chilies & 30.8 \\
dried & 30.6 \\
beer & 30.3 \\
pancetta & 30.0 \\
mustard & 29.8 \\
spice & 28.4 \\
sliced & 28.3 \\
cinnamon & 28.0 \\
warm & 27.8 \\
\bottomrule
\end{tabular}
        \caption{\centering{\textbf{\color{speech_better}ASR Better (\aucdelta)}}}
        \label{tab:speech_better}
    \end{subtable} %
    \begin{subtable}[b]{0.145\linewidth}
        \centering
        \tiny
\begin{tabular}[b]{lc}
\toprule
sandwich & -18.0 \\
stove & -15.4 \\
tuna & -14.3 \\
again & -12.6 \\
mince & -11.2 \\
wok & -8.9 \\
burger & -8.8 \\
pizza & -8.4 \\
serve & -7.8 \\
4 & -6.6 \\
mussels & -6.6 \\
tray & -6.3 \\
bowl & -5.9 \\
\bottomrule
\end{tabular}
        \caption{\centering{\textbf{\color{vid_better} Video Better (\aucdelta)}}}
        \label{tab:video_better}
    \end{subtable} %
    \caption{Per-word classification results using ASR and/or Video features. Each point in the scatterplot represents a different word-type; x-coordinate values show how well a word is predicted by ASR-token features; y-coordinate values show how well a word is predicted by video features. Tables (a)-(d) show word types that are easy, universally difficult, better-predicted-by-ASR, and better-predicted-by-video, respectively.}
    \label{fig:per_word_performance}
\end{figure*}

\section{Complementarity of Video and ASR} 
We now turn to the question of \emph{why} multimodal models produce better captions: what type of signal does video contain that speech does not (and vice versa)? 
Our initial idea was to quantitatively compare the captions generated by AT versus AT+Video; however, because the dataset is relatively small, we were unable to make observations about the generated captions that were statistically significant.\footnote{In general, making concrete statements about the causal link between inputs and outputs of sequence-to-sequence models is challenging, even in the text-to-text case, see \newcite{alvarez2017causal}.}

Instead, we examine properties of the ASR-token--based and visual features directly.
Following a procedure inspired from \cite{lu2008high,berg2012understanding,dai2018neural,mahajan2018exploring}, we consider the auxiliary task of predicting presence/absence of unigrams in the ground truth captions from features extracted from corresponding segments. We train two unimodal classifiers, one using ASR-token--based features and one using visual features, and measure their relative capacity to predict different word types; the goal is to measure which word types are most-predictable from the ASR tokens and, conversely, which ones are most-predictable from the visual features.

For each segment, we predict the unigram distribution of its corresponding caption using a unimodal softmax classifier: for simplicity, we use a 2-layer, residual deep averaging network \cite{iyyer2015deep} for both the visual and ASR-based classifier.
We measure per-word-type performance using \auc, which is word-frequency independent.

Specifically --- for each word type $w$ (e.g., $w=\texttt{beer}$) we measure how well $w$ is predicted by the classifier based on ASR / spoken tokens 
\aucspeech (e.g., \aucspeechw{beer}= 98) and, conversely, how well $w$ is predicted by the visual classifier \aucvideo (\aucvideow{beer}= 68).
For a given word type, we measure its overall difficulty by averaging \aucspeech and \aucvideo; we call this \aucmean (\aucmeanw{beer}= 83).
Similarly, we measure the difference in difficulty by subtracting \aucspeech and \aucvideo to give \aucdelta (\aucdeltaw{beer}= 30) with higher values indicating that a word type is predicted better by the spoken-token features compared to the visual features.
We plot \aucspeech versus \aucvideo for 382 words in Fig.~\ref{fig:per_word_performance} (results are averaged over 10 cross-val splits).

\mparagraph{Absolute Performance}
Points in the upper-right quadrant of Fig.~\ref{fig:per_word_performance} represent words that are easy for both visual and ASR-token--based features to predict, whereas points in the lower-left represent words that are more difficult.
Specific ingredients, e.g., ``nori" and ``mozzarella," are often easy to detect, as are actions closely associated with particular objects (e.g., ``dough" is almost always the object being ``knead"-ed).
Conversely, pronouns (e.g., ``it") and conjunctions (e.g., ``or") are universally difficult to predict. %

\mparagraph{Visual vs. ASR-token--based features}
In general, ASR-token--based features carry greater predictive power, as evidenced by the skew towards the bottom right in the scatterplot in Fig.~\ref{fig:per_word_performance}.
One pattern in the cases where speech features perform better (Fig. \ref{tab:speech_better}) is that words are often modifiers, e.g., \emph{white} (pepper), \emph{sea} (salt), \emph{dried} (chilies), \emph{olive} (oil), etc.
Indeed, small, detailed distinctions may be often difficult to make from visual features, e.g., ``vegetable oil" and ``olive oil" may look identical in most YouTube videos.

Nonetheless, there are types better predicted by video features (Fig. \ref{tab:video_better}).
Often, these are cases that require unstated, background knowledge, i.e., references to objects not explicitly stated by the speaker(s).
To quantify this observation, for each word type we compute the likelihood that it is \emph{stated} by the speaker in the video, given that it appears in the ground-truth caption, i.e.,
$P(w \in \texttt{ASR}~|~w \in \texttt{GT})$.
Aside from trivial cases (e.g., words misspelled in the GT never appear in the ASR), words that are often unstated include action words (e.g., ``place", ``crush") and cookware (e.g., ``pan", ``wok", ``pot").
Words that are often stated include specific ingredients (e.g., ``honey", ``coconut", ``ginger").
In contrast to word frequency (which is uncorrelated with \aucdelta, Spearman $\rho \approx 0$), stated rate \emph{is} correlated with \aucdelta ($\rho = 0.44$, $p<.01$).

\section{Oracle Object Detection}
The results in Table~\ref{tab:main_results} indicate that, while adding visual information yields statistically significant improvements to the ASR-only model, the improvements are not large in magnitude.
This leaves open the question of whether (a) any visual information simply does not provide much additional information on top of ASR, or (b) we need better visual modeling.
We take a first step in addressing this question by experimenting with an ``oracle'' object detector that provides perfect-precision predictions.\footnote{High-precision object detectors are gaining popularity in the computer vision community because the training data is easier to annotate, e.g., \newcite{OpenImages}.}
If even oracle object detection does not help, then the answer is more likely (a) rather than (b) above.

As part of a YouCook2 data release, bounding box annotations for selected objects in the recipe text \cite{ZhLoCoBMVC18} were provided. 
Unfortunately, while these could have served as an oracle, the actual annotations are only available for a small fraction of the data.
Instead, we consider the set of 62 object labels made available.
We simulate a high-precision, oracle object detector by identifying -- per video segment -- the overlap between (morphology-normalized) groundtruth caption mentions and the 62 object labels available.\footnote{This oracle is unlikely to be achievable, as it assumes 100\% precision
for the 62 objects considered (which also implies modeling {\em which} objects to talk about, a non-trivial task in itself \cite{berg2012understanding}).
}
For instance, for the groundtruth caption ``put the mushrooms in the pan", the oracle object detector yields ``mushroom" and ``pan". 89\% of segments receive at least one oracle object.
The oracle object detections are then fed into the Transformer encoder (in random order), either by themselves (Oracle) or along with the ASR token sequence (AT+Oracle).
We perform the same cross-validation experiments as described in \S\ref{sec:experiments}, and report the average \rouge (we observe similar trends with other metrics):

{\footnotesize
\begin{center}
\begin{tabular}{c | c c c c}
     &  AT & AT+Video & Oracle & AT+Oracle \\ \midrule
{\small ROUGE-L} & 35.5 & 36.7 & 40.8 & \textbf{45.5} \\
\end{tabular}
\end{center}}
Because the AT+Oracle model achieves large improvements over AT+Video, we suspect that building higher-quality visual representations is a promising avenue for future work.

\begin{figure}
    \centering
    \includegraphics[width=0.6\linewidth]{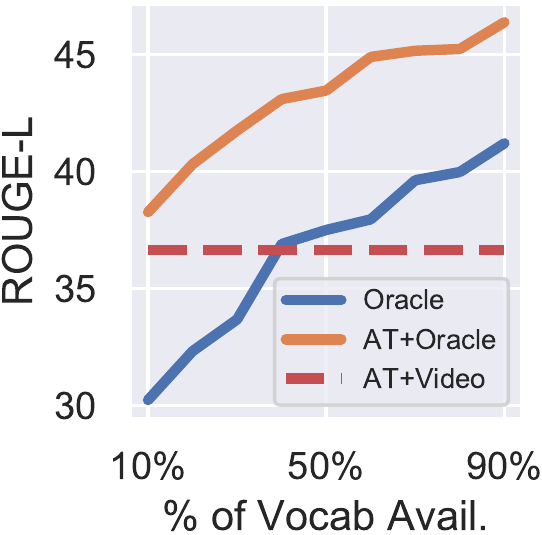}
    \caption{The performance of the oracle methods increases as they are given access to an increasing number of object types.}
    \label{fig:oracle_sweep}
\end{figure}

How weak of an oracle can still produce high performance? Fig.~\ref{fig:oracle_sweep} shows performances of models using \emph{subsets} of the 62 objects (most frequent 10\% of objects through 90\%) over one cross-validation fold. AT+Oracle gives better performance than AT+Video by detecting \emph{just 6 object types,} and the oracle by-itself (which is only given access to object sets) achieves comparable performance to AT+Video with 30 object types.
These results suggest that, at least for this task, the Transformer decoder is likely not the main performance bottleneck, as it is able to paste-together unordered object detections into captions effectively.

\section{Conclusion}

In this work, we demonstrate the impact of incorporating both visual and ASR-token--based features into instructional video captioning models.
Additional experiments investigate the complementarity of the visual and speech signals.

Our oracle experiments suggest that performance bottlenecks likely derive from the input encoding, as the decoder is able to paste-together even simple sets of object detections into high-quality captions. Future work would thus be well-suited to investigate better models of input data.
Given the small size of the dataset, transfer learning may prove fruitful, e.g., pre-training the encoder with an unsupervised, auxiliary task; work contemporaneous with our submission from the computer vision community suggests that transfer learning indeed is a promising direction \cite{sun2019videobert,sun2019contrastive,miech19howto100m}.

\pagebreak
\mparagraph{Acknowledgements} We would like to thank Maria Antoniak, Nan Ding, Sebastian Goodman, Jean Griffin, Fernando Pereira, Chen Sun, and the anonymous reviewers for their helpful comments.

\bibliography{refs}
\bibliographystyle{acl_natbib}

\end{document}